# Equivalence and Synthesis of Causal Models*

TS Verma[†]

< verma@cs.ucla.edu >

Cognitive Systems Laboratory
Computer Science Department
University of California
Los Angeles, CA 90024

Judea Pearl

< judea@cs.ucla.edu >

Cognitive Systems Laboratory
Computer Science Department
University of California
Los Angeles, CA 90024

June 18, 1990

## Abstract

Scientists often use directed acyclic graphs (dags) to model the qualitative structure of causal theories, allowing the parameters to be estimated from observational data. Two causal models are equivalent if there is no expirement which could distinguish one from the other. A canonical representation for causal models is presented which yields an efficient graphical criterion for deciding equivalence, and provides a theoretical basis for extracting causal structures from empirical data. This representation is then extended to the more general case of an embedded causal model, that is, a dag in which only a subset of the variables are observable. The canonical representation presented here yields an efficient algorithm for determining when two embedded causal models are equivalent, and leads to a model theoretic definition of causation in terms of statistical dependencies.

## 1 Introduction

The use of dags as a language for describing causal models has been popular in the behavioral sciences [Blalock 71], [Duncan 75] and [Wright 34], decision analysis [Howard and Matheson 81], [Olmsted 84] and [Shachter 85] and evidential reasoning [Pearl 88], and has also received extensive theoretical studies [Geiger and Pearl 89], [Geiger and Verma 90], [Glymour et al 1987], [Pearl and Verma 87], [Shachter 85], [Smith 89], [Spirtes et al 90] and

*This work was supported, in part, by NSF grant IRI-88-2144 and NRL grant N000-89-J-2007.

[†]Supported by an IBM graduate fellowship.

[Verma and Pearl 90]. One problem that has arisen in the course of these studies is that of non-uniqueness; it is quite common for two different causal models to be experimentally indistinguishable, hence, equally predictive. Formally, let a *causal theory* be a pair $T = < D, \Theta >$, where $D$ is a dag, called the *causal model* of $T$, and $\Theta$ a set of parameters compatible with $D$ (i.e., sufficient for forming a probability distribution for which $D$ is a Bayesian network). We say that two causal models $D_1$ and $D_2$ are equivalent if for every theory $T_1 = < D_1, \Theta_1 >$ there is a theory $T_2 = < D_2, \Theta_2 >$ such that $T_1$ and $T_2$ describe the same probability distribution, and vice versa.

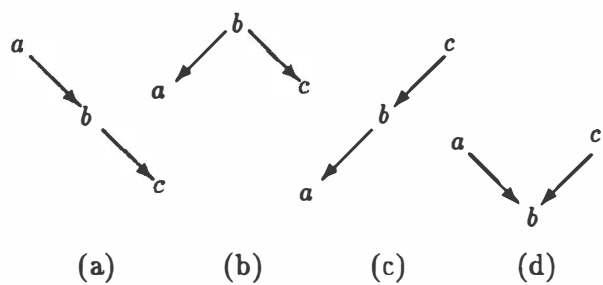

(a)     (b)     (c)     (d)

Figure 1: Three of the four models are equivalent.

For example, consider the four causal models of Figure 1. The parameters required for the first model are $P(a)$, $P(b|a)$ and $P(c|b)$. The second requires estimations for $P(b)$, $P(a|b)$ and $P(c|b)$. It is easy to see that these two models are equivalent since by Bayes law, $P(a)P(b|a) = P(ab) = P(b)P(a|b)$, hence the values obtained for the first set of parameters completely determine the values of the second, and vice versa. Similarly, the third model is equivalent to

the first two since its parameters, $P(c)$, $P(b|c)$ and $P(a|b)$ can be determined from either of the first two sets. However, the fourth model is quite different; its parameters are $P(a)$, $P(c)$ and $P(b|ac)$ which cannot be determined from any of the previous sets.

The fact that the first three models are equivalent to each other but not the fourth is easily seen in terms of the independence information conveyed by the corresponding dags. The first three all represent the independence statement $I(a, b, c)$ which is read "$a$ is independent, given $b$, of $c$", whereas the fourth represents the statement $I(a, \emptyset, c)$, which is read "$a$ is marginally independent of $c$". It is known that the statistical meaning of any causal model can be described economically by its stratified protocol, which is a list of independence statements that completely characterize the model [Geiger and Pearl 89], [Pearl and Verma 87] and [Verma and Pearl 90]. Furthermore, any independence statement that logically follows from the stratified protocol can be graphically determined in linear time via the *d-separation* criterion [Geiger and Verma 90] and [Geiger et al 89]. Thus, the question of equivalence of causal models reduces to the question of equivalence of protocols: two dags are equivalent if and only if each dag's protocol holds in the other [Pearl et al 89]. This solution is both intuitive and efficient. However, it has two drawbacks; it is difficult to process visually and it does not generalize to embedded causal models.

Embedded causal models are useful for modeling theories that cannot be modeled via simple dags. For example, if there are unobserved variables which cause spurious correlations between the observable variables it may be necessary to embed the observables in a larger dag containing "hidden" variables in order to build an accurate model. Even when there exists a simple causal model that fits theory, it might be desirable to embed the model in a larger dag to satisfy some higher level constraints. For example, suppose that every causal model that fits a given set of data contains the link $a \rightarrow b$. Furthermore, suppose that $b$ occurs before $a$ in time and that causality is assumed to be temporal. Under these circumstances, the simple causal models are inconsistent with the higher level constraints on the temporal direction of causality; one way of avoiding this conflict is to hypothesize the existence of an unknown common cause, i.e. $a \leftarrow \alpha \rightarrow b$. See Figures 3 and 4 for examples of the use of hidden variables.

This paper is organized as follows. Section 2 provides an efficient criterion for deciding the equiva-lence of two models, and a canonical representation called a *pattern* for describing the class of all models equivalent to a given dag. Section 3 extends this construction to the case of embedded causal models. Theorems will be stated without proofs, a full detail of which can be found in [Verma 90]. In section 4, the Theorems of the previous two sections are applied to the problem of recovery of a causal model from statistical data.

## 2   Patterns of Causal Models

It is not difficult to observe that equivalent dags have common features. For example, two dags that represent equivalent causal models must have the same adjacency structure. Two nodes of a dag are adjacent, written $\overline{ab}$ if either $a \rightarrow b$ or $a \leftarrow b$. That adjacency is invariant among equivalent dags follows from Lemma 1 which describes the principle relationship between adjacency and unseparability [1] (parts 1 and 2) as well as the relationships between separability and d-separation [2] given two particular special sets of nodes in the dags (parts 3 and 4). Let the ancestor set $A_{ab}$ of a pair of variables $a$ and $b$ be defined as the union of the sets of ancestors of $a$ and $b$ (less $ab$), and similarly, the parent set $P_{ab}$ of the pair be defined as the union of the sets of parents of $a$ and $b$ (less $ab$).

**Lemma 1** *Let $a$ and $b$ be two nodes of a dag $D$; the following four conditions are equivalent:*

*(1) $a$ and $b$ are adjacent in $D$*
*(2) $a$ and $b$ are unseparable in $D$*
*(3) $a$ and $b$ are not d-separated by $A_{ab}$ in $D$*
*(4) $a$ and $b$ are not d-separated by $P_{ab}$ in $D$*

**Proof:** (Sketch) That (1) implies (2) follows from the fact that a link is a path which cannot be deactivated; and (2) trivially implies (3) since unseparability means the lack of d-separation in any context, including $A_{ab}$. Since every path activated by $P_{ab}$ is also activated by $A_{ab}$, it follows that (3) implies (4). The final implication, that (4) implies (1) follows from the observation that if $a$ and $b$ are not d-separated given $P_{ab}$, then there must be an active path between them. If this path contains a node, other than $a$ or $b$, it would have to contain at least one head-to-head node since it is active given the parents of $a$ and $b$; and for the same reason, the head-to-head node nearest to $a$ on the path would be a descendant of $a$, similarly the one nearest $b$ would be a descendant of $b$. Both of these head-to-head nodes would have to be in or be an ancestor of a node in $P_{ab}$ for the path to be active, but the one nearest $a$ could not be an ancestor of $a$, hence both it and $a$ would be ancestors of $b$. Similarly, both $b$ and the head to head node nearest it would have to be ancestors of $a$, but this would imply the existence of a directed loop, hence the path cannot contain any nodes other than $a$ and $b$. Therefore the nodes are adjacent. □

---

[1] two variables are unseparable just in case there is no set that d-separates them.
[2] the predicate $I_D(\cdot)$ denotes d-separation in the dag $D$.

The major consequence of this lemma is that adjacency is a property determined solely by d-separation, hence remains invariant among equivalent dags.

A set of equivalent dags possesses another important invariant property, namely the directionality of the uncoupled head-to-head links (i.e. $a \rightarrow b \leftarrow c$ are uncoupled if $a$ and $c$ are not adjacent). There are other links whose directionality remains invariant, but these can easily be determined from the uncoupled head-to-head links. The following lemma summarizes this important class of links with invariant directionality.

**Lemma 2** *If the nodes $a$, $c$, $b$ form the chain $\overline{acb}$ while $a$ and $b$ are not adjacent, then $c$ is head-to-head between $a$ and $b$ if and only if $a$ and $b$ are not separable by any set containing $c$. That is, for any dag $D$, $\overline{acb} \in D$ and $\overline{ab} \notin D \Rightarrow$*

$$[a \rightarrow c \leftarrow b \in D \iff \neg I_D(a, Sc, b) \; \forall_{S \subseteq U - abc}]$$

The proof of this lemma relies upon the inherent differences between a head-to-head junction and the other types of junctions (tail-to-tail and head-to-tail). The major ramification of Lemma 2 is that the directionality of a certain class of links can be determined from d-separation alone. The implications this may have on the prospects of inferring causal relationships from independence statements are briefly discussed in section 4 and in detail in [Verma 90].

Together, these lemmas form a necessary and sufficient condition for equivalence, previously stated in [Pearl et al 89]:

**Theorem 1** *Two dags are equivalent if and only if they have the same links and same uncoupled head-to-head nodes.*

The proof of this theorem is based on the lemmas along with an inductive step showing that every active path in one dag has a corresponding active path in the other. The importance of Theorem 1 is that the equivalence of two causal models can be determined by a simple graphical criterion.

Since the two invariant properties of a dag identified in the lemmas are a sufficient condition for equivalence, they lead to a natural canonical representation of its equivalent class. Simply construct a *partially directed graph* by removing the arrowheads from any link of the dag that is not identified by Lemma 2. This partially-directed graph will be called the *rudimentary pattern* of the causal model. Since the rudimentary pattern can be defined solely in terms of d-separation, it follows that each equivalence class of causal models has a unique pattern; hence, two causal models are equivalent if and only if they have the same pattern. This is a useful view of the problem since the patterns can be constructed efficiently [3].

Lemma 2 only identifies some of the invariant arrowheads of a causal model, but since identification of this class is sufficient for deciding equivalence, it follows that the remainder of the invariant arrowheads are completely determined by this class. It is not difficult to identify the remainder of the invariant arrowheads as some of the undirected links of a rudimentary pattern cannot be arbitrarily directed without either (1) creating a new uncoupled head-to-head node or (2) creating a directed loop. Since these undirected links are essentially constrained to a certain direction, it is desirable to define a *completed pattern* in which they are directed as constrained. The completed pattern reflects each and every invariant arrow head. Furthermore, both rudimentary patterns and completed patterns offer a compact summary of each and every dag in an equivalence class.

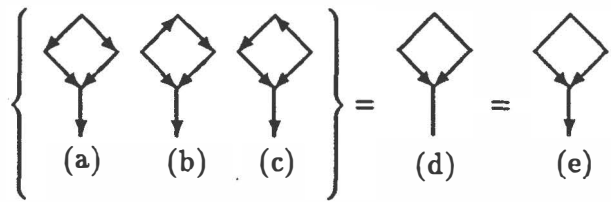

$$\left\{ \quad \text{(a)} \quad \text{(b)} \quad \text{(c)} \quad \right\} = \text{(d)} = \text{(e)}$$

Figure 2: Equivalence class of models.

---

[3] Note that comparison of patterns is polynomial since the nodes are labeled

For example, in Figure 2, the rudimentary pattern (d) and the completed pattern (e) each summarizes the dags in the equivalence class { (a), (b), (c) }. Any extension of either pattern into a full dag that does not create new uncoupled head-to-head nodes will be a dag in the equivalence class. There are three such extensions in the example of Figure 2.

# 3  Embedded Causal Models

Partially-directed graphs offer an excellent tool for describing the equivalence classes of causal models; it would be desirable to find a similar structure for embedded causal models. Such a structure requires the ability to represent a direct non-causal correlation between two variables. In a simple dag, whenever two variables are unseparable, there must be a directed link between them, dictating that either the first causes the second or the second causes the first. There is no way to represent the existence of an unknown common cause, as illustrated in the following embedded causal model (Figure 3 (a)). Assume $a$, $b$, $c$ and $d$ are the observables and $\alpha$ is unobservable. There is no dag that can represent the dependencies between $a$, $b$, $c$ and $d$ using these variables only. However, the *hybrid graph* (Figure 3 (b)) which contains a *bi-directional* link does represent these dependencies. (Under a natural extension of d-separation [Verma 90]).

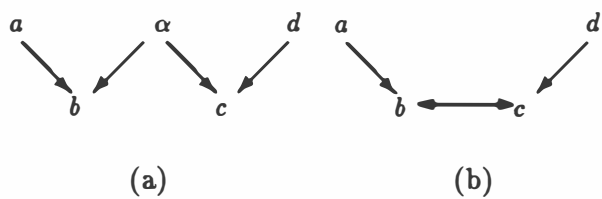

(a)                    (b)

Figure 3: The representation of a hidden common cause.

For hybrid graphs, the notation $\overrightarrow{ab}$ denotes the existence of a link with at least an arrow head pointing at $b$, namely either $a \rightarrow b$ or $a \leftrightarrow b$, while $\overline{ab}$ denotes the existence of a link without any constraints on its orientation. Thus, for example, when applied to a dag, $\overline{ab}$ means $a \rightarrow b$ or $a \leftarrow b$; while in hybrid graphs $\overline{ab}$ denotes the existence of any of the four possible types of links, (namely, $a - b$, $a \rightarrow b$, $a \leftarrow b$ and $a \leftrightarrow b$). Hybrid graphs can be used to represent *patterns* of embedded causal models according to the following definition.

**Definition 1** *(Embedded Pattern) Given a dag $D$ over the variables $U_D$, of which $U_O \subseteq U_D$ are observable, the rudimentary pattern $P$ of $D$ restricted to $U_O$ is defined as the hybrid graph with fewest arrowheads that satisfies the following conditions:*

*(1)* $\overline{ab} \in P \iff \neg I_D(a, S, b) \;\; \forall S \subseteq U_O - ab$

*(2)* $\overrightarrow{ab}$ *if* $\exists c \in U_O$ *such that:* $\;\; \overline{abc} \in P$, $\overline{ac} \notin P$ *and* $\neg I_D(a, Sb, c) \;\; \forall S \subseteq U_O - abc$

Rudimentary embedded patterns can be extended into completed embedded patterns (or simply, embedded patterns) in much the same way that simple patterns are completed. The same constraints can be used for the completion, namely, no arrow head can be added to the pattern that would (1) create a new uncoupled head-to-head node or (2) create a strictly directed cycle. However, note that a strictly directed cycle contains only singly directed arrows.

While this defines a unique pattern for embedded every dag, it does so in terms of d-separation conditions over subsets all of $U_O$, which, in principle, might require an exponential number of tests. The next two lemmas show that patterns can be formed in polynomial time. Lemma 3 delineates the relationship between adjacency in the pattern and unseparability in the causal model (parts 1 and 2) and provides a practical criterion for determining separability in terms of a simple d-separation test (part 3) and a graphical test (part 4). The graphical test is defined in terms of an *inducing path*:

**Definition 2** *(Inducing Path) An inducing path between the variables $a$ and $b$ of an embedded causal model is any path $\rho$ satisfying the following two conditions:*

*(1) Every observable node on $\rho$ is head-to-head on $\rho$.*
*(2) Every head-to-head node on $\rho$ is in $A_{ab}$.*

**Lemma 3** *Let $P$ be the pattern of a dag $D$ with respect to the observables $U_O \subset U_D$ and $a, b \in U_O$ be two observables; the following statements are equivalent:*

*(1) $a$ and $b$ are adjacent in $P$*
*(2) $a$ and $b$ are unseparable in $D$ (over $U_O$)*
*(3) $a$ and $b$ are not d-separated by $A_{ab} \cap U_O$ in $D$*
*(4) $a$ and $b$ are connected by an inducing path in $D$*

**Proof:** (Sketch) By definition, (1) is equivalent to (2) and (2) implies (3). To show that $\neg I(a, A_{ab} \cap U_O, b)$ implies the existence of an inducing path, consider that this dependency implies the existence of a path $\rho$, between $a$ and $b$ which is active given $A_{ab} \cap U_O$. Since $A_{ab} \cap U_O$ only contains ancestors of $a$ and $b$ it follows that every head-to-head node on $\rho$ must be in $A_{ab}$. Thus any observable node on $\rho$ that is not head-to-head would be in $A_{ab} \cap U_O$ and would serve to deactivate the path, so every observable node on $\rho$ must be head-to-head. Therefore $\rho$ is an inducing path.

To show that the existence of an inducing path implies unseparability relative to $U_O$ hence finish the proof, consider any two nodes $a$ and $b$ which are connected by an inducing path $\rho$. To show $a$ and $b$ are not d-separated in any context of $U_O$, consider any context $S$ which deactivates $\rho$ (if $\rho$ is active for every context, then the two nodes are unseparable). Since the only observable nodes of $\rho$ are head-to-head, only head-to-head nodes could serve to deactivate $\rho$. Each head-to-head node on $\rho$ must be in $A_{ab}$ and at least one must be inactive, given $S$ (otherwise the path would be active given $S$). If all inactive head-to-head nodes are ancestors of $a$ then consider the one closest to $b$, call it $y$. The portion of $\rho$ between $y$ and $b$ is active, and the ancestry path from $y$ to $a$ can be added to form an active path between $a$ and $b$ given $S$. On the other hand, if any of the inactive head-to-head nodes is ancestor of $b$ then pick the head-to-head ancestor of $b$ which is closest to $a$ on $\rho$ and call it $x$. Every inactive head-to-head node between $a$ and $x$ must be an ancestor of $a$ (if any exist), hence there must be an active path between $a$ and $x$ (either the portion of $\rho$ between $A$ and $X$, or the ancestry path from the head-to-head node between $a$ and $x$ which is closest to $x$ concatenated with the portion of $\rho$ from that node to $x$). Since $x$ is an ancestor of $b$, the ancestry path from $x$ to $b$ can be concatenated to the path from $a$ to $x$ to form an active path between $a$ and $b$ given $S$. Thus $A$ and $B$ are unseparable. □

Lemma 3 describes how links are induced in $P$ by paths of $D$. The next lemma will describe how to determine the directionality of these links in terms of the inducing paths.

**Lemma 4** *For any pattern $P$, $\overrightarrow{ab}$ if and only if there is a node $c$ adjacent to $b$ but not to $a$ (in $P$) such that both edges $\overline{ab}$ and $\overline{bc}$ were induced by paths (of $D$) which ended pointing at $b$.*

Lemmas 3 and 4 provide a polynomial time algo-rithm for constructing the characteristic pattern of any embedded causal model. The final theorem completes the original task of deciding equivalence.

**Theorem 2** *Two embedded causal models are equivalent if and only if they have the same pattern.*

Thus, Theorem 2 gives validity to the notion of a pattern as a characteristic representation of an embedded causal model. An interesting consequence of this theorem is given by the following corollary:

**Corollary 1** *There are fewer than $5^{|U_O|^2}$ distinct embedded causal models containing $|U_O|$ variables; moreover, every embedded causal model is equivalent to a simple dag with fewer than $|U_O|^2$ variables.*

Part 1 follows from the fact that every embedded causal model is equivalent to its pattern, and every pattern contains fewer than $|U_O|$ edges (there are four types of edges). The second part stems from the fact that a bi-directional link $a \leftrightarrow b$ in a pattern can be represented by a single hidden common cause $\alpha$ of the observable variables, namely, $a \leftarrow \alpha \rightarrow b$.

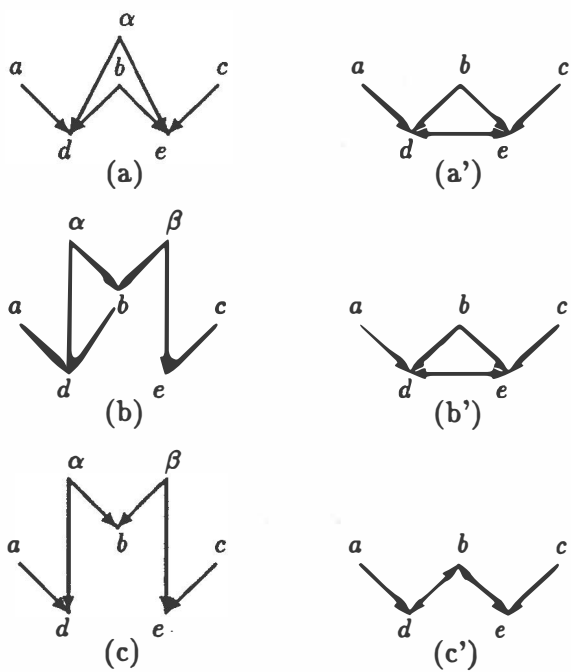

Figure 4: The patterns reveal which two models are equivalent.

Figure 4 contains three embedded causal models (a), (b) and (c) over the observable variables

$\{a, b, c, d, e\}$ as well as their completed patterns (a'), (b') and (c') respectively. The patterns indicate that the first two causal models are equivalent to each other but not to the third; while $a$ and $b$ are marginally independent in (e) they are dependent in both (a) and (b). Figure 4 (b) demonstrates that a hidden common cause is not equivalent to a bidirectional link since it is important to recognize the paths they may induce.

# 4 Applications to the Synthesis of Causal Models

The problem of deciding the equivalence of (embedded) causal models is fundamental to causal reasoning and theory building, as it allows us to determine which structural properties of the model (e.g. connectivity or directionality) can be substantiated by data and which serve merely for representational convenience. The canonical representations presented in this paper offer an efficient solution to this problem since they can be constructed (from the causal models) in polynomial time. They can also be used to solve the broader problem of model subsumption [Verma 90].

The construction of these canonical representations is based on (conditional) independence relationships, thus suggesting the possibility of extracting causal models directly from statistical information. Such application meets with the difficulty that, in general, probability distributions do not define unique graphical models. In other words, given that the data is generated by some causal theory $T = < D, \Theta >$, it is always possible to contrive the parameters $\Theta$ to yield spurious independencies, not shown in $D$, that fit another theory $T' = < D', \Theta' >$, with $D'$ not equivalent to $D$. [Spirtes et al 90] show that, under some reasonable assumptions, the occurrence of such spurious independencies is a rare event of measure zero, and therefore argue that it is natural in causal modeling to assume that the underlying distribution is dag-isomorphic [4], albeit allowing for the inclusion of unobserved variables.

Under the assumption that the observed distribution is dag isomorphic, Theorem 1 permits the recovery of the underlying structure uniquely, modulo the equivalence class defined by its pattern. One such re-covery algorithm is proposed in [Spirtes et al 90] and several alternatives are discussed in the sequel.

The basic algorithm has three parts; the first part is an application of Lemma 1 that identifies the links of the pattern. The second part of the algorithm is an application of Lemma 2 which adds directionality to some of the links, thus forming the rudimentary pattern. The final part of the algorithm consists of completing the rudimentary pattern into a full pattern (if desired).

## Recovery Algorithm

1. For each pair of variables $a$ and $b$, search for a separating set $S_{ab}$, (i.e. such that $I(a, S_{ab}, b)$ holds.

   If there is no such $S_{ab}$, place an undirected link between the variables.

2. For each pair of non-adjacent variables $a$ and $b$ with a common neighbor $c$, test the statement $I(a, cS_{ab}, b)$.

   If the statement holds then continue.
   If the statement is false, add arrowheads at $c$, (i.e. $a \rightarrow c \leftarrow b$).

3. Complete the pattern.

The complexity of this algorithm is bounded by the first step, which by brute force would require an exponential search for the set $S_{ab}$. It can be greatly reduced by the generation of a Markov network. A Markov network is the undirected graph formed by linking every pair of variables $a$ and $b$ that are dependent given the rest of the variables (i.e. $\neg I(a, U - ab, b)$). The Markov network of a dag-isomorphic distribution has the property that the parents of any variable in the dag form a clique in the network. Since Lemma 1 states that any two variables $a$ and $b$ are separable if and only if they are separated by their parent set $P_{ab}$, the search for a separating set can be confined to the cliques that contain a or b. Thus, the complexity is bounded, exponentially, by the size of the largest clique in the Markov network, and this coincides with the theoretical lower bound for recovery of a dag from independence information [Verma 90].

One drawback of the Markov network reduction is that it is not applicable to embedded causal models because it rests on part (4) of Lemma 1; no parallel lemma exists for embedded models. However, the basic algorithm stated above, by virtue of resting on Theorem 2 can be used to recover embedded causal

---

[4] A probabilistic distribution is dag-isomorphic permitting all its dependencies and independencies to be displayed in some dag

model as well. The only difference is in the output; when the algorithm is applied to a dag isomorphic distribution, every link is guaranteed to be assigned at most one arrowhead (a particular arrowhead may actually be assigned multiple times, but no link will receive an arrowhead on both ends). However, when the distribution is isomorphic to an embedded dag it is possible for a link to be assigned an arrowhead on both ends, hence the recovery of a bi-directional link.

The invariant nature of the arrows in a pattern can form the basis for a general non-temporal definition of causation; one that determines the direction of causal influences from statistical data without resorting to chronological information, and one that applies to general distributions, including those that are not isomorphic to embedded dags. The essence of this definition can be articulated by taking as *models* of our theory the set $\mathcal{P}$ of all patterns that are consistent with, an observed distribution, namely, patterns that are minimal I-maps of the distribution.

**Definition 3 (Genuine and Potential Cause)** *c is a genuine cause of e if c causes e in every consistent model (i.e. every pattern of $\mathcal{P}$ contains the directed arrow $c \rightarrow e$). c is a potential cause of e if c causes e in some consistent model (i.e. some pattern of $\mathcal{P}$ contains $c \rightarrow e$) and e never causes c in any consistent model (i.e. no pattern of $\mathcal{P}$ contains $c \leftarrow e$).*

The vertical arrow in Figure 2 (e) is an example of a genuine cause, since this arrow cannot be emulated by a hidden common cause of the two end points (in any consistent embedded model). The other arrows in Figure 2 (e) represent potential causes when viewed in the context of embedded models, because each can be represented by a common hidden cause in some equivalent causal model.

Since the number of patterns over $|U|$ variables is finite, Definition 3 is operational. However, the existence of an effective algorithm which can determine causation by means other than enumerating the patterns of $\mathcal{P}$ is an open question. If the observed distribution is isomorphic to an embedded dag, then $\mathcal{P}$ contains only one unique pattern; that which is generated by the recovery algorithm. This pattern contains all the information required for identifying the genuine and potential causes [Verma 90]. However, when applied to general distributions the arrows assigned in the generated pattern may or may not coincide with the model-theoretic definition of genuine and potential causes.

[Spirtes et al 90] have proposed an algorithm for identifying causal relationships which accepts many, but not all, of the genuine and potential causes in distributions that are isomorphic to embedded dags. The relationships identified by [Spirtes et al 90] correspond to the singly directed arrows of the rudimentary pattern.

In practice, every recovery algorithm must face the problem of inferring independence relations from sampled data. The number of samples required to reliably test the assertion $I(a, S_{ab}, b)$ grows exponentially with the size of $S_{ab}$. A reasonable approximating algorithm for recovering a dag (or embedded dag) could be devised based upon the following redefinition of the independence relation:

**Definition 4 (Reliable Independence)** *$I(a, S, b)$ holds reliably whenever the set of hypotheses $\{P(a|S) = P(a|Sb)\}$ is confirmed for each instantiation of S for which a sufficient number of samples are available to reliably test the hypothesis.*

This notion of reliable independence is captured by taking as a measure of dependency the (conditional) sample cross entropy [Pearl 88, page 392]:

$$\tilde{H}(a, b|S) \stackrel{\text{def}}{=} \sum_{a,b,S} \hat{P}(a, b, S) \log \frac{\hat{P}(a, b|S)}{\hat{P}(a|S)\hat{P}(b|S)}$$

where $\hat{P}$ stands for the sample frequency and the summation ranges over all instantiations of $a$, $b$ and $S$. We see that terms involving small samples (i.e., low values of $\hat{P}(a, b, S)$) are automatically discounted relative to those of larger samples.

One issue that has not been addressed is that of deterministic nodes, such as those representing functional dependencies among variables. These nodes cannot be completely represented by the causal models considered in this paper, as they require a refinement of d-separation studied in [Geiger et al 89] and [Pearl et al 89]. The issues introduced by deterministic nodes are discussed in [Verma 90].

## Acknowledgement

The problem of deciding equivalence of embedded causal models was posed by Clark Glymour and communicated to us by Dan Geiger.

# References

[Blalock 71] H.M. Blalock, *Causal Models in The Social Sciences*. Macmillian, London, 1971.

[Duncan 75] O.D. Duncan, *Introduction to Structural Equation Models*. Academic Press, New York, 1975.

[Geiger and Pearl 89] D. Geiger and J. Pearl, Logical and Algorithmic Properties of Independence and Their Application to Bayesian Networks, UCLA Cognitive Systems Laboratory, *Technical Report CSD-890035 (R-123)*, February 1989. To appear in *Annals of Mathematics and AI*, Special Issue on Statistics and AI.

[Geiger et al 89] D. Geiger, T.S. Verma and J. Pearl, d–Separation: From Theorems to Algorithms, *Proceedings*, 5th Workshop on Uncertainty in AI, Windsor, Ontario, Canada, August 1989, pp. 118-124.

[Geiger and Verma 90] D. Geiger and T.S. Verma, Identifying Independence in Bayesian Networks, UCLA Cognitive Systems Laboratory, *Technical Report CSD-890028 (R-116)*, To appear in *Networks*, John Wiley and Sons, Sussex, England, 1990.

[Glymour et al 1987] C. Glymour, R. Scheines, P. Spirtes and K. Kelly. *Discovering causal structure*. Academic Press, New York, 1987.

[Howard and Matheson 81] R.A. Howard and J.E. Matheson, Influence Diagrams, chapter 8, in *The Principles and Applications of Decision Analysis*, Vol. II, Strategic Decisions Group, Menlo Park, California, 1981.

[Olmsted 84] S.M. Olmsted, On Representing and Solving Decision Problems, Ph.D. Thesis, Engineering-Economic Systems Dept., Stanford University, Stanford California, 1984.

[Pearl et al 89] J. Pearl, D. Geiger and T.S. Verma, The Logic of Influence Diagrams, in R.M. Oliver and J.Q. Smith (Eds), *Influence Diagrams, Belief Networks and Decision Analysis*, John Wiley and Sons, Ltd., Sussex, England 1990. A shorter version, in *Kybernetica*, Vol. 25:2, 1989, pp. 33-44.

[Pearl 88] J. Pearl, *Probabilistic Reasoning in Intelligent Systems: Networks of Plausible Inference*. Morgan Kaufmann Publishers, Inc, San Mateo, California, 1988.

[Pearl and Verma 87] J. Pearl and T.S. Verma, The Logic of Representing Dependencies by Directed Graphs, *Proceedings*, AAAI Conference, Seattle, WA. July, 1987, pp. 374-379.

[Shachter 85] R.D. Shachter, Evaluating Influence Diagrams, in A.P. Basu (Eds), *Reliability and Quality Control*, Elsevier, 1985, pp. 321-344.

[Spirtes et al 90] P. Spirtes, C. Glymour and R. Scheines, Causality from Probability, in G. Mc-Kee, ed., *Evolving Knowledge in Natural and Artificial Intelligence*, Pitman, 1990.

[Smith 89] J.Q. Smith, Influence Diagrams for Statistical Modeling, *The Annals of Statistics*, Vol. 17(2):654-672, 1989.

[Verma 90] T.S. Verma, Invariant Properties of Causal Models. In preparation.

[Verma and Pearl 90] T.S. Verma and J. Pearl, Causal Networks: Semantics and Expressiveness, UCLA Cognitive Systems Laboratory, *Technical Report 870032 (R-65)*, June 1986. Also in *Uncertainty in AI 4*, R. Shachter, T.S. Levitt and L.N. Kanal (eds), Elsevier Science Publishers, 1990.

[Wright 34] S. Wright, The Method of Path Coefficients. *Ann. Math. Statistics* 5:161-215, 1934.